\documentclass[10pt,twocolumn,letterpaper]{article}

\usepackage{cvpr}              

\definecolor{cvprblue}{rgb}{0.21,0.49,0.74}
\usepackage{graphicx}
\usepackage{booktabs}
\usepackage{amsmath}
\usepackage{amssymb}
\usepackage{textcomp}
\usepackage[utf8]{inputenc}
\usepackage{kotex}
\usepackage{multirow}
\usepackage{indentfirst}
\usepackage{makecell} 
\usepackage{wrapfig}
\usepackage{lipsum}
\usepackage[accsupp]{axessibility}  
\usepackage[pagebackref,breaklinks,colorlinks,allcolors=cvprblue]{hyperref}

\def\paperID{4164} 
\def\confName{CVPR}
\def\confYear{2025}

\title{Gyro-based Neural Single Image Deblurring}

\author{Heemin Yang$^1$
\quad
Jaesung Rim$^2$
\quad
Seungyong Lee$^1$
\quad
Seung-Hwan Baek$^{1,2}$
\quad
Sunghyun Cho$^{1,2}$ \vspace{-2mm}\\ \\
POSTECH CSE$^1$ \& GSAI$^2$\\
{\tt\small \{heeminid, jsrim123, leesy, shwbaek, s.cho\}@postech.ac.kr}
}

\begin{document}
\def\MethodName{GyroDeblurNet}
\def\SynthDataName{GyroBlur-Synth}
\def\RealDataName{GyroBlur-Real}
\def\DataName{GyroBlur}
\def\GyroReprName{camera motion field}
\newcommand{\XX}{{\textcolor{red}{XX}}}

\newcommand{\Eq}[1]   {Eq.\ (#1)}
\newcommand{\Eqs}[1]  {Eqs.\ (#1)}
\newcommand{\Fig}[1]  {Fig.\ #1}
\newcommand{\Figs}[1] {Figs.\ #1}
\newcommand{\Tbl}[1]  {Table #1}
\newcommand{\Tbls}[1] {Tables #1}
\newcommand{\Sec}[1]  {Section #1}
\newcommand{\SSec}[1] {Section #1}
\newcommand{\Secs}[1] {Sections #1}
\newcommand{\Alg}[1]  {Algorithm #1}

\definecolor{rjscolor}{rgb}{0.15, 0.26, 0.75}
\newcommand{\sunghyun}[1]{{\textcolor[rgb]{0.6,0.0,0.6}{[sunghyun: #1]}}}
\newcommand{\sh}[1]{{\textcolor[rgb]{0.2,0.6,0.6}{[BAEK: #1]}}}
\newcommand{\old}[1]{{\textcolor{lightgray}{#1}}}
\newcommand{\heemin}[1]{{\textcolor{cyan}{[heemin: #1]}}}
\newcommand{\rjs}[1]{{\textcolor[rgb]{0.15, 0.26, 0.75}{rjs: #1}}}
\newcommand{\sean}[1]{{\textcolor{blue}{[Seungyong: #1]}}}

\newcommand{\Q}[1]{{\textcolor{red}{Question: #1}}}
\newcommand{\Todo}[1]{{\textcolor{red}{Todo: #1}}}

\newcommand\T{\rule{0pt}{2.3ex}}       
\newcommand\B{\rule[-1.0ex]{0pt}{0pt}} 

\renewcommand{\paragraph}[1]{\vspace{2pt}\noindent\textbf{#1}~~}
\maketitle
\begin{abstract}

In this paper, we present \MethodName{}, a novel single-image deblurring method that utilizes a gyro sensor to resolve the ill-posedness of image deblurring.
The gyro sensor provides valuable information about camera motion that can improve deblurring quality.
However, exploiting real-world gyro data is challenging due to errors from various sources.
To handle these errors, \MethodName{} is equipped with two novel neural network blocks: a gyro refinement block and a gyro deblurring block.
The gyro refinement block refines the erroneous gyro data using the blur information from the input image.
The gyro deblurring block removes blur from the input image using the refined gyro data and further compensates for gyro error by leveraging the blur information from the input image.
For training a neural network with erroneous gyro data, we propose a training strategy based on the curriculum learning.
We also introduce a novel gyro data embedding scheme to represent real-world intricate camera shakes.
Finally, we present both synthetic and real-world datasets for training and evaluating gyro-based single image deblurring.
Our experiments demonstrate that our approach achieves state-of-the-art deblurring quality by effectively utilizing erroneous gyro data.
\end{abstract}

\vspace{-4mm}    
\section{Introduction}
\label{sec:intro}

Blur caused by camera shakes severely degrades image quality as well as the performance of various computer vision tasks.
To overcome the image degradation caused by blur, single image deblurring has been extensively studied for decades~\cite{cho2009fast, hirsch2011fast, zhang2013non, pan2014deblurring, whyte10nonuniform, joshi2010gyro, hu2016gyro}.
Recently, various deep neural network (DNN)-based approaches have been proposed and have shown significant performance improvements over classical approaches.~\cite{nah2017gopro, zhang2019dmphn, tao2018srn, kupyn2018deblurgan, kupyn2019deblurganv2, zamir2021mprnet, cho2021mimounet, chen2021hinet}.
Nonetheless, they still fail on images with large blur due to the severe ill-posedness of deblurring.

To address the ill-posedness of image deblurring, several attempts have been made to exploit the gyro sensors that most smartphones are now equipped with. 
Classical approaches based on blur models use gyro data to guide blur kernel estimation~\cite{joshi2010gyro, sindelar2013inertial, sindelar2014inertial, park2014gyro, hu2016gyro, mustaniemi2018fastgyro}. 
However, their performance is limited by restrictive blur models that struggle to handle noise, nonlinear camera response functions, saturated pixels, moving objects, and other challenges effectively.
With the advancement of deep learning, DNN-based approaches have also been proposed to exploit gyro data~\cite{mustaniemi2019deepgyro, lee2020gyro, ji2021eggnet, ren2024informer, varghese2024gyro}. 
Thanks to the remarkable capabilities of DNNs and the rich information from gyro sensors, these methods demonstrate superior performance compared to the classical gyro-based approaches.

\begin{figure}[t]
\includegraphics[width=\linewidth]{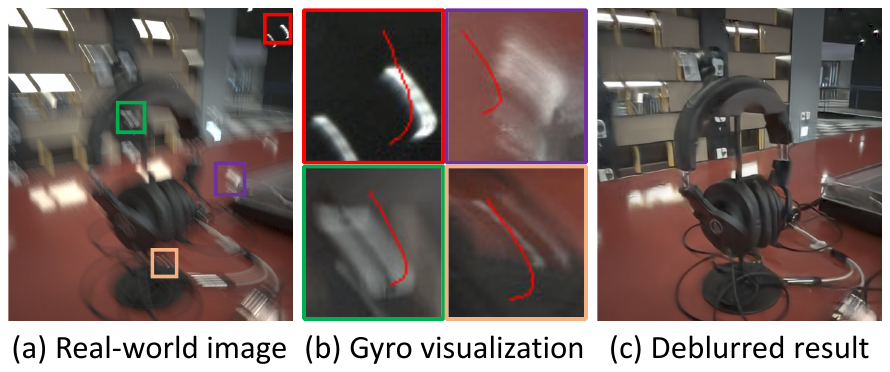}
\centering
\caption{
Our method shows robust performance even with errors in real-world gyro data. (a) A real-world blurry image. (b) Gyro data visualization onto the blurry image. The gyro data do not perfectly match the blur trajectories due to errors. (c) Our deblurred result.
}
\vspace{-5mm}
\label{fig:gyro_error_example}
\end{figure}

However, recent DNN-based methods that utilize gyro data are still limited in handling real-world blurred images. 
First, gyro data from mobile devices such as smartphones usually contain a significant amount of noise caused by various sources~\cite{kim2013,Fedasyuk2020,joshi2010gyro}. 
Second, gyro sensors measure the angular velocity around their center. 
Therefore, recent gyro-based methods assume that the gyro sensor and the camera share the same center and that the camera is only affected by rotational motions. 
However, in practice, the positions of the camera and the gyro sensor differ, and the camera experiences more complex shakes. 
Third, real-world blurred images may have moving objects with different blur trajectories that cannot be captured by gyro sensors. 
All these factors cause the motion information encoded in the gyro data to be inconsistent with the blur in the blurred image (\cref{fig:gyro_error_example}-(b)), making it challenging to exploit gyro data in real-world image deblurring.
We refer to such inconsistency between the gyro data and blur as \emph{gyro error}.

In this paper, we propose \emph{\MethodName}, a novel gyro-based single image deblurring approach that produces high-quality deblurring results even in the presence of gyro error (\cref{fig:gyro_error_example}-(c)). 
To address gyro error, \MethodName{} adopts a carefully designed network architecture that includes novel gyro refinement blocks and gyro deblurring blocks. 
The gyro refinement block corrects gyro data errors using the blur information from the input image. 
On the other hand, the gyro deblurring block removes blur from the input image using the refined gyro data. 
Additionally, to further compensate for gyro error, the gyro deblurring block leverages both the blur information in the input image and the input gyro data. 
Despite these blocks, training a neural network with erroneous gyro data is challenging, as the network might be trained to ignore the input gyro data and rely solely on the input image. 
To resolve this issue, we propose a curriculum-learning-based training strategy~\cite{bengio2009curriculum}.

To handle complex real-world camera shakes, we also present the \emph{\GyroReprName{}}, a novel gyro data embedding scheme that can represent complex camera motions.
Real-world blurry images often exhibit complex blur trajectories.
Nonetheless, recent gyro-based approaches~\cite{mustaniemi2019deepgyro, lee2020gyro, ji2021eggnet, ren2024informer} assume smooth camera motions and represent camera motions using one or two motion vectors per pixel or a couple of homographies for the entire image, which leads to low-quality deblurring results.
To resolve this issue, our \GyroReprName{} is designed to represent complex camera motions using multiple vectors per pixel while maintaining a comparable memory footprint exploiting the spatially smooth nature of camera shakes.

Previous methods~\cite{mustaniemi2019deepgyro, lee2020gyro, ji2021eggnet, ren2024informer} rely on synthetic datasets to train and evaluate the methods.
However, their gyro data do not reflect real-world camera shakes that occur when capturing still-shot images either.
Gyro data of previous methods were generated by random sampling~\cite{zhang2020dataset} or obtained from the Visual-Inertial dataset~\cite{schubert2018imu}.
However, the Visual-Inertial dataset was originally developed for visual odometry and SLAM, so its camera motions were collected from constantly moving cameras, which differ from camera shakes of still-shot images.

For training and evaluating gyro-based deblurring for real-world blurred images,
we propose two datasets: \emph{\SynthDataName} and \emph{\RealDataName}.
\SynthDataName{} is a synthetic dataset for both training and evaluation.
The dataset consists of sharp and blurred image pairs with their corresponding gyro data.
The gyro data are acquired from a smartphone device to reflect real-world camera shakes.
The blurred images are synthetically generated using the acquired gyro data and the blur synthesis process of RSBlur~\cite{rim2022rsblur} for realistic blurred images.
\RealDataName{} is a real-world blur dataset for qualitative evaluation with no ground-truth sharp images, and it provides real-world blurred images paired with gyro data collected from a smartphone.

We validate the performance of \MethodName{} on both synthetic and real-world datasets.
Our experiments demonstrate that \MethodName{} clearly outperforms state-of-the-art deblurring methods both quantitatively and qualitatively.
Our contributions can be summarized as follows:

\begin{itemize}
    \item We propose a novel gyro-based single image deblurring approach, {\em \MethodName{}}. Extensive quantitative and qualitative evaluations show that \MethodName{} outperforms existing deblurring methods.
    \item To handle gyro error, we introduce a carefully designed network equipped with novel gyro refinement and gyro deblurring blocks.
    We also develop a novel gyro data embedding to represent complex camera shakes and propose a curriculum learning-based training scheme.
    \item We present synthetic and real image-gyro paired datasets with realistic camera motions, \SynthDataName{} and \RealDataName{}, for training and evaluating gyro-based single image deblurring methods.
\end{itemize}
\section{Related Work}
\label{sec:related work}

A variety of DNN-based single image deblurring approaches have recently been proposed~\cite{nah2017gopro, zhang2019dmphn, tao2018srn, kupyn2018deblurgan, kupyn2019deblurganv2, zamir2021mprnet, cho2021mimounet, chen2021hinet, wang2022uformer, Tsai2022Stripformer, kong2023fftformer}.
To push the limit of single image deblurring, they introduce various network architectures and training schemes.
Nevertheless, single image deblurring is still a challenging problem due to its high ill-posedness, and they still fail on images with large blur.

To resolve the ill-posedness of deblurring, there have been attempts to leverage inertial measurement sensors such as the gyro sensor and accelerometer as they provide valuable information on the camera motion.
Joshi \etal \cite{joshi2010gyro} and Hu \etal \cite{hu2016gyro} exploit the gyro sensor and accelerometer to estimate spatially-varying blur kernels.
On the other hand, Park and Levoy~\cite{park2014gyro} and Mustaniemi \etal \cite{mustaniemi2018fastgyro} use only the gyro sensor to estimate blur kernels, because blur kernel estimation using accelerometer measurements requires estimating the depth of a scene, the device orientation to compensate the gravity effect, and the initial velocity of the device, which is often error-prone in practice.
Moreover, rotational motion can be assumed as the dominant factor of camera shake blur as discussed by Whyte \etal~\cite{whyte10nonuniform}.

In the deep learning era, there have been a few works that exploit gyro data.
Mustaniemi \etal~\cite{mustaniemi2019deepgyro} propose a DNN-based approach that uses gyro data for the first time.
They introduce a U-Net-based architecture~\cite{ronneberger2015unet} that takes a concatenation of a blurred image and the gyro data as input, which is also adopted by Lee \etal~\cite{lee2020gyro}.
Ji \etal~\cite{ji2021eggnet} adopt deformable convolutions to adjust convolution filters based on gyro data.
Ren \etal~\cite{ren2024informer} propose a transformer-based network that utilizes gyro data.
However, these methods assume accurate gyro data, neglecting potential errors.
Moreover, they assume smooth camera motions and model camera motions using a vector field consisting of either one~\cite{mustaniemi2019deepgyro, ren2024informer} or two vectors~\cite{ji2021eggnet} per pixel, or a couple of homographies~\cite{lee2020gyro} representing the beginning and end position of the camera during the exposure time.
Unfortunately, such restrictive assumptions do not hold in real-world cases as discussed in \cref{sec:intro}.


Deblurring using gyro data can be framed as non-blind deconvolution, with gyro data serving as blur kernels. 
However, most non-blind deconvolution approaches assume accurate blur kernels without errors, making them unsuitable for erroneous gyro data~\cite{richardson1972rldeconv, wiener1949wienerdeconv, levin2007deconv, krishnan2009deconv}. Some attempts have been made to address kernel errors in non-blind deconvolution. 
Vasu \etal~\cite{vasu2018nonblinderror} apply varying regularization strengths in a non-blind deblurring algorithm and use a neural network to combine results and refine artifacts induced by kernel errors. 
Nan \etal~\cite{nan2020nonblinderror} tackle kernel error artifacts by estimating residual errors induced by kernel errors using a CNN. 
However, these approaches assume mild errors in kernels and fail to handle the significant errors presented in gyro data, as will be demonstrated in our experiments.


\begin{figure}[t]
\includegraphics[width=\linewidth]{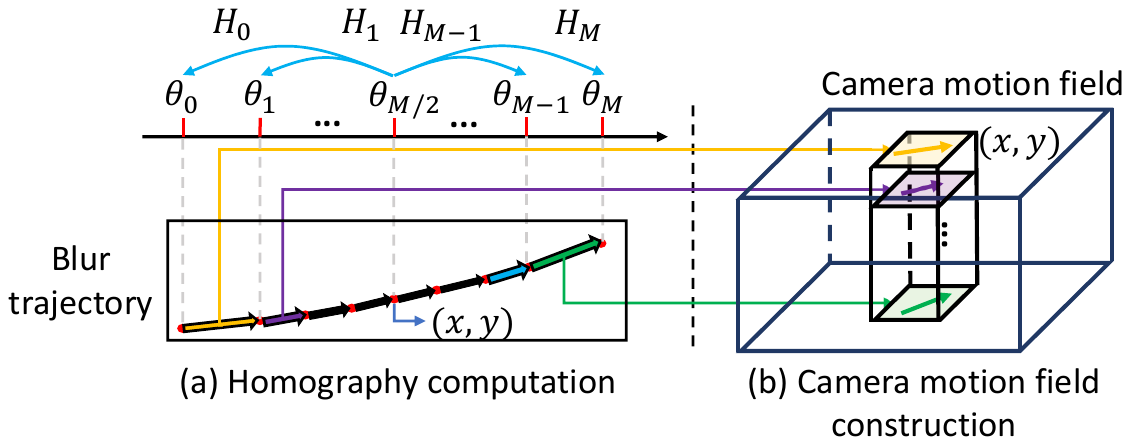}
\centering
\vspace{-5mm}
\caption{Camera motion field construction.
(a) Computing homography and warped pixel coordinates.
(b) Constructing \GyroReprName{} by stacking motion vectors.}
\vspace{-1mm}
\label{fig:dmgv_generation}
\end{figure}
\section{\MethodName}
\label{sec:method}

The goal of \MethodName{} is to estimate a sharp deblurred image $D$ from an input blurred image $B$ and its corresponding gyro data $G$ possibly with large errors.
Specifically, we define $G$ as a sequence of gyro data such that $G=\left\{ g_0, \cdots, g_{T-1}\right\}$,
where $g_t$ is the $t$-th gyro data consisting of three angular velocities during the exposure time.
$T$ is the number of gyro data, which is proportional to the exposure time.
Instead of directly using $G$, we devise a novel gyro data embedding scheme named \GyroReprName{} to effectively handle complex camera shakes of an arbitrary exposure duration.
In the rest of this section, we explain the gyro data embedding scheme, network architecture, and training strategy of \MethodName{} in detail.


\begin{figure}[t]
\centering
\includegraphics[width=\linewidth]{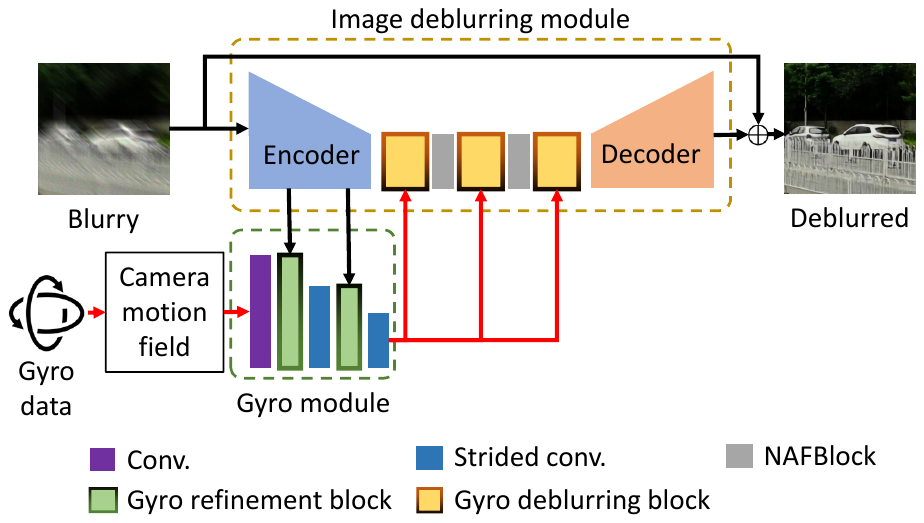}
\caption{Network architecture of \MethodName{}.
}
\vspace{-1mm}
\label{fig:network_arch}
\end{figure}

\begin{figure*}[t]
\includegraphics[width=\linewidth]{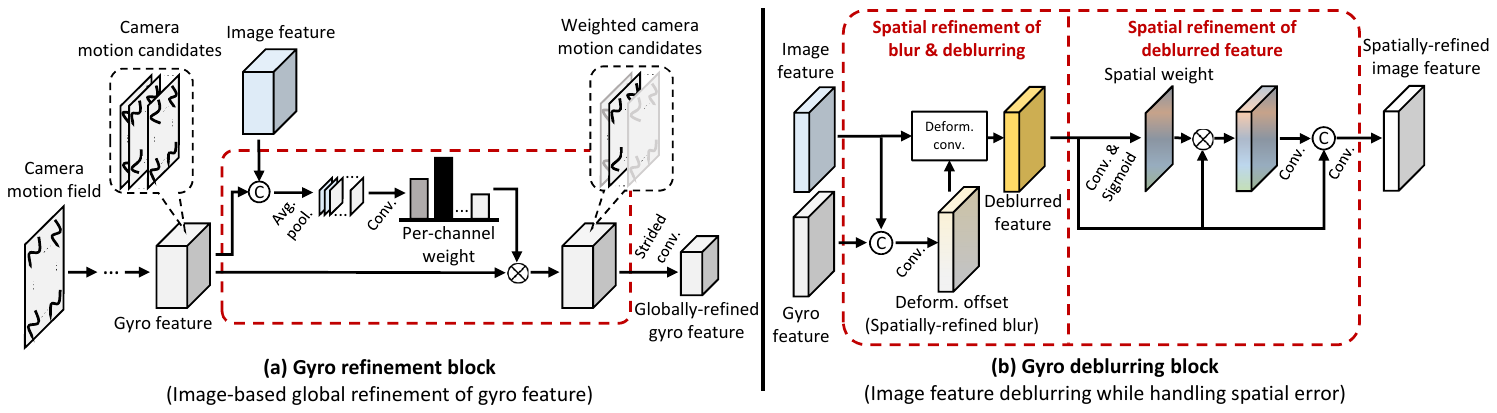}
\centering
\caption{Detailed architectures of the modules in the \MethodName{}. (a) Architecture of the gyro refinement block. (b) Architecture of the gyro deblurring block.
}
\vspace{-1mm}
\label{fig:module_arch}
\end{figure*}


\subsection{Gyro Data Embedding}
\label{method:gyro_data_embedding}

\MethodName{} first converts input gyro data sequence $G$ into a \GyroReprName{} $\mathcal{V}$ to handle temporally complex camera shakes of an arbitrary length $T$.
The conversion process is illustrated in \cref{fig:dmgv_generation}.
A \GyroReprName{} has a fixed channel size $2M$ where $M$ is a hyperparameter representing the number of vectors in \GyroReprName{}, regardless of the length of the input gyro data to allow it to be fed to convolutional neural networks.
To achieve this, we first resample the input gyro data sequence $G$ of an arbitrary length $T$ to $M+1$ samples using cubic-spline interpolation.
We denote the resampled sequence as $G'=\left\{ g'_0, \cdots, g'_{M}\right\}$ where $g'_m$ is a resampled gyro sample.
The hyperparameter $M$ is set to an even number so that we have the same number of gyro data before and after the temporal center. 

Then, assuming that the camera and the gyro shares the same center and there exists neither off-center rotational nor translational motions, we integrate the angular velocities and obtain $M+1$ camera orientations $\theta_m$, where $m=0,\cdots,M$, including the camera orientations at the beginning and end of the exposure.
Note that, despite our restrictive assumption on the camera motion, \MethodName{} can successfully perform deblurring thanks to its robustness to gyro errors.
For each $\theta_m$, we compute a homography $H_m$ as $H_m = KR(\theta_m)K^{-1}$ 
where $K$ is the camera intrinsic matrix and $R(\theta_m)$ is the rotation matrix corresponding to $\theta_m$.
To prevent unnecessary shift after deblurring, we chose the temporal center as a reference by assuming $\theta_{M/2}$ at the temporal center to be $(0,0,0)$ so that $H_{M/2}$ is an identity matrix when computing $\theta_m$.

Finally, from the obtained homographies, we compute a \GyroReprName{} $\mathcal{V}$, which illustrates how each pixel is blurred.
Specifically, we define a \GyroReprName{} $\mathcal{V}$ as a tensor of size $W/{s}\times H/{s} \times 2M$ where $W$ and $H$ are the width and height of the input blurred image, and $s$ is a scaling factor.
Then, at each position $(x,y)$ in $\mathcal{V}$,
we compute the warped coordinates of $(x,y)$ by $H_m$ for all $m$ and compute the difference between temporally consecutive coordinates to obtain $M$ vectors.
By stacking the vectors for all the spatial positions, we construct $\mathcal{V}$.
In our experiments, we set $M=8$ and $s = 2$.
\cref{fig:gyro_error_example}-(b) visualizes camera motion field vectors onto a real-world blurry image.

Note that similar representations for gyro data based on vector fields have also been proposed by previous approaches~\cite{mustaniemi2019deepgyro,ji2021eggnet}, as discussed in \cref{sec:related work}.
However, unlike these methods that rely on only one or two vectors to represent camera shakes at each pixel, our \GyroReprName{} offers a more sophisticated representation.
By adjusting the hyperparameter $M$, it can effectively capture temporally intricate camera shake patterns. 
Furthermore, our approach leverages the spatial smoothness of camera shakes, utilizing the hyperparameter $s$. 
As a result, it can provide a detailed representation of complex motions while maintaining a comparable memory footprint.

\subsection{Network Architecture}
\label{sec:network_arch}

\MethodName{} takes a blurred image $B$ and a \GyroReprName{} $\mathcal{V}$ as input and estimates a deblurred image $D$. 
The network consists of two modules: an image deblurring module and a gyro module (\cref{fig:network_arch}). 
The image deblurring module deblurs a blurred image $B$ with the aid of gyro data. 
It then produces a residual $R$ that is added back to $B$ to produce $D$. 
The image deblurring module adopts a U-Net architecture~\cite{ronneberger2015unet} and the NAFBlock~\cite{chen2021hinet} as its basic building block.
The image deblurring module adopts gyro deblurring blocks in its bottleneck to perform deblurring using gyro features from the gyro network.

On the other hand, the gyro module takes a \GyroReprName{} $\mathcal{V}$ and refines it with the aid of image features from the image deblurring module.
The gyro module consists of a convolution layer to embed an input \GyroReprName{} $\mathcal{V}$ into the feature space, gyro refinement blocks and strided convolution layers.
The gyro refinement blocks refine gyro features with the aid of image features from the encoder of the image deblurring module.
The gyro module progressively refines gyro features through gyro refinement blocks and strided convolution layers, and obtains a gyro feature of the same spatial size as the feature maps of the bottleneck in the image deblurring module.
In the following, the gyro refinement block and the gyro deblurring block, which are the two key components of \MethodName{}, are explained in detail.
For a detailed network architecture including feature dimensions, refer to the supplementary material.

\paragraph{Gyro refinement block} 
\cref{fig:module_arch}-(a) illustrates the structure of the gyro refinement block.
The camera motion information in the input \GyroReprName{} $\mathcal{V}$ may contain significant global errors caused by gyro sensor noise, unknown rotational centers, and missing translational motions, as visualized in \cref{fig:gyro_error_example}-(b). 
The gyro refinement block is designed to compensate for these global errors using the blur information in the input image based on the following intuition.

The gyro refinement block takes a gyro feature computed from the input \GyroReprName{} $\mathcal{V}$ as input. 
This input gyro feature encodes multiple camera motion candidates across different channels, obtained by introducing perturbations to the input camera motion information in the previous layers.
To globally refine the gyro feature, the gyro refinement block adaptively selects the channels of the gyro feature that are consistent with the blur information in the input image. 
Specifically, the gyro refinement block takes an image feature from the image deblurring module as an additional input and concatenates it with the input gyro feature. 
It then computes channel-wise weights~\cite{hu2018channelattention} through global average pooling and $1\times1$ convolution layers, and multiplies them to the gyro feature. 
Following the gyro refinement block, a strided convolution layer further refines the gyro feature while reducing its spatial resolution.



\paragraph{Gyro deblurring block} 
The gyro deblurring block, illustrated in \cref{fig:module_arch}-(b), performs deblurring of the image feature using the gyro feature from the gyro module. 
The input image $B$ may have spatially-variant blur caused by camera shakes and moving objects. 
To effectively remove such blur, the gyro deblurring block consists of two sub-blocks: the first sub-block performs spatial refinement of the blur information and deblurring of the image feature using the refined blur information, while the second sub-block further refines the deblurred image feature.

Specifically, the first sub-block concatenates the image feature and gyro feature, computing spatially-refined blur information in the form of deformable convolution kernel offsets. 
It then performs deblurring of the image feature using these offsets. 
The second sub-block further refines the deblurred image feature using spatial attentions~\cite{xu2021spatialattentiondeblurring}. 
Spatial attention weights are computed from the deblurred image block through convolution and sigmoid layers, and multiplied with the deblurred image feature, which is further processed through convolution and concatenation layers.

\subsection{Training Strategy}
\label{subsec:training_strategy}
To address the challenge of training a deblurring network with erroneous gyro data, we propose a curriculum learning-based training strategy. 
This approach, inspired by its success in handling noisy data in prior work~\cite{bengio2009curriculum, sangineto2018self, shi2016weakly, sinha2020curriculum}, aims to progressively guide the network in leveraging gyro information effectively despite inaccuracies.

Our training strategy initially trains the network with error-free gyro data, and gradually increases error in the training data.
To this end, we first assume that, for each blurred image $B$, its corresponding error-free gyro data $G_{GT}$ is given as well as its ground-truth sharp image $S_{GT}$.
From $G_{GT}$, we compute a noise-free \GyroReprName{} $\mathcal{V}_{clean}$ and a noisy \GyroReprName{} $\mathcal{V}_{noisy}$.
We generate $\mathcal{V}_{noisy}$ by randomly perturbing the rotational center and adding noise to the gyro data $G_{GT}$.
We then compute a blended \GyroReprName{} $\mathcal{V}_{\alpha}=(1-\alpha)\cdot\mathcal{V}_{clean}+\alpha\cdot\mathcal{V}_{noisy}$ where $\alpha$ is a blending parameter.
We gradually increase $\alpha$ from 0 to 1 during training.
For more details including the noisy \GyroReprName{} generation and the scheduling of $\alpha$, refer to the supplementary material.

\section{\DataName~Dataset}
\label{sec:dataset}


\subsection{\SynthDataName}

\SynthDataName{} provides 14,600 and 640 synthetically blurred images of size $720\times1280$ for the training and test sets.
The dataset also provides ground-truth sharp images and the corresponding gyro data.
To faithfully mimic real-world blurred images, the dataset covers various degradation sources, such as moving objects, noise, and saturation.

To build a dataset with realistic camera motions, we implemented an Android application that records gyro data on a mobile phone.
We used a Samsung Galaxy S22 smartphone for collecting gyro data with the sampling rate of 200 Hz.
We collected two sequences of gyro data for generating training and test datasets with realistic hand-shake motions.
The gyro data sequences are recorded for 195 sec.~and 60 sec., and have 39,116 and 12,065 samples, respectively.
We then used the gyro data to synthesize blurry images.
To this end, we used sharp frames of the 4KRD dataset~\cite{deng20214krd}.
To generate a blurry image, we first randomly sampled a sequence of $10$ consecutive gyro data corresponding to the exposure time of $1/20$ seconds.
We then interpolated them by eight times to synthesize a continuous blur trajectory, and computed homographies from the interpolated gyro data.
Finally, we warped the sharp image using the homographies, and blended them to obtain a blurry image.

To create synthetic blurred images with moving objects that have different blur trajectories, we used a simple method.
We select a single instance from the COCO dataset~\cite{lin2014coco} and assume it moves at a constant speed in a blurry image.
We randomly choose its position and the direction and speed of motion.
We then add the object to the warped sharp images before blending them during the blur synthesis process described earlier.

To reflect real-world noise and saturated pixels, we also adopt the blur synthesis pipeline of RSBlur~\cite{rim2022rsblur}.
Specifically, we synthesize blur in the linear sRGB space following the aforementioned process.
Then, we synthesize saturated pixels, convert the image into the camera RAW space, synthesize noise, and convert the noisy RAW image to the linear sRGB image.
We use the camera parameters of a Samsung Galaxy S22 ultra-wide camera, and the noise distribution estimated from it.
We refer the readers to \cite{rim2022rsblur} for the detailed blur synthesis process.

To generate noisy \GyroReprName{}s, we add gyro errors with gyro data noise and randomly perturbed rotation centers.
To obtain realistic gyro noise, we measured the gyro noise distribution from a stationary smartphone (Samsung Galaxy S22) placed on a table. 
Gyro noise is modeled as a Gaussian distribution for each rotation axis independently.
For more details on the dataset, we refer the readers to the supplementary material.

Our synthetic dataset generation requires minimal device-specific input, brief gyro sequences, and simple camera calibration, significantly reducing manual data collection. We capture sufficient gyro data within minutes and obtain the necessary camera parameters for the RSBlur pipeline from a few calibration images. While we do not explicitly model the gyro’s position relative to the camera, this is implicitly addressed as gyro errors such as rotational center error. Our experiments demonstrate effective blur removal regardless of these errors.



\if{0}
\sunghyun{===========}

To simulate a realistic blurry image with continuous blur trajectory, we interpolated gyro data by eight times, simulating 1600 Hz sampling speed.
\sunghyun{아래 있는 문장은 무슨 의미인지 이해 안 감. 그리고 gyro 몇개씩 샘플링하는지? 즉, 노출 시간을 어떻게 가정했는지? random sampling하는지?}

\heemin{
x-axis에 대한 noise $\sim$ $\mathcal{N}(-0.00005643153, 0.0008631607^2)$ rad/s
}

\heemin{
y-axis에 대한 noise $\sim$ $\mathcal{N}(-0.00006369004, 0.0015023947^2)$ rad/s
}

\heemin{
z-axis에 대한 noise $\sim$ $\mathcal{N}( 0.00021379517, 0.0007655643^2)$ rad/s
}

Gyro data are sampled from the interpolated gyro data from which homography can be estimated for each sample of the measurement.
The homography $H$ can be computed with the following equation~\cite{mustaniemi2019deepgyro, lee2020gyro, ji2021eggnet}
\begin{equation}
\label{eq:image_homography}
    H = KRK^{-1}
\end{equation}
where $K$ denotes the camera intrinsic matrix and $R$ denotes the rotation matrix which is computed from gyro data.

\sunghyun{object motion blur는 여태까지 한번도 언급 없다가 여기서 갑자기 언급되는데 굉장히 뜬금없이 들린다.}
\heemin{Introduction에서 gyro sensor의 limitation 언급할 때 언급하는 게 좋아 보입니다. 수정해 보도록 하겠습니다.}
To generate blurry images, we use sharp frames of the 4KRD dataset~\cite{deng20214krd}.
The sharp frames are warped by applying the homography $H$ which is computed using \cref{eq:image_homography}.
In order to simulate object motion blur, we superimpose instances of COCO dataset~\cite{lin2014coco} onto the sharp frames.
Object motions are generated by randomly sampling direction, location and magnitude of the object motion blur each time generating a blurry image.
The warped images are then averaged to generate a blurry image and are center-cropped to remove dark area that occurs due to image warping.
\heemin{RSBlur 설명 추가}
In order to simulate realistic-looking blurry images, we apply blur synthesis pipeline of RSBlur~\cite{rim2022rsblur}.
We use the camera ISO parameters of Samsung Galaxy S22 ultra-wide camera to apply the pipeline.

\SynthDataName{} has 14,600 and 640 blurry images of $720\times1280$ resolution for the training and test sets.
The dataset also includes additional raw data such as the original sequences of gyro measurements and camera intrinsic parameters.
For more details on the dataset such as detailed generation procedure and ISO parameters that are used in RSBlur synthesis pipeline, we refer the readers to the supplementary material.

\fi

\begin{table*}[t]
\centering
\scalebox{0.84}{
\begin{tabular}{clcccccccc}
\toprule[1.5pt]
\multirow{2}{*}{Method}    & \multirow{2}{*}{Models}                 & \multicolumn{2}{c}{\SynthDataName} & & \multicolumn{2}{c}{\RealDataName{}-S}   & \multirow{2}{*}{Param. (M)} & \multirow{2}{*}{MACs (G)} & \multirow{2}{*}{Time (s)} \\ \cline{3-4}\cline{6-7}
 & & PSNR\ $\uparrow$ & SSIM\ $\uparrow$ & & NIQE\ $\downarrow$ & TOPIQ\ $\uparrow$ & & & \\ \midrule[1.5pt]
\multirow{6}{*}{\centering\parbox{2cm}{\centering Single-image\\deblurring}}                           & 
                             MPRNet~\cite{zamir2021mprnet}            & 25.03 & 0.7081 & & 5.25 & 0.397 & 20.13  & 10927 & 0.998  \\
                           & NAFNet~\cite{chen2022nafnet}             & 25.06 & 0.7085 & & 5.27 & 0.409 & 17.11  & 227    & 0.106    \\
                           & AdaRevD-B (NAFNet)~\cite{mao2024adarevd} & 25.61 & 0.7281 & & 5.18 & 0.425 & 43.81  & 2658   & 1.661    \\
                           & Uformer-B~\cite{wang2022uformer}         & 25.72 & 0.7334 & & 4.80 & 0.431 & 50.88  & 2143   & 1.061    \\
                           & Stripformer~\cite{Tsai2022Stripformer}   & 25.93 & 0.7398 & & 4.71 & 0.456 & 19.71  & 2397   & 1.367    \\ 
                           & FFTformer~\cite{kong2023fftformer}       & 26.01 & 0.7481 & & 4.98 & 0.434 & 16.56  & 1894   & 1.879    \\ \midrule
Non-blind                  & Vasu \etal~\cite{vasu2018nonblinderror}  & 19.63 & 0.4526 & & 6.13 & 0.317 & 10.41  & -                                 & 160.176 \\ 
deblurring                 & Nan \etal~\cite{nan2020nonblinderror}    & 22.22 & 0.5313 & & 5.81 & 0.348 & 26.16  & 
                           41170    & 27.069 \\ \midrule
 \multirow{4}{*}{\centering\parbox{2cm}{\centering Gyro-based\\deblurring}}         & DeepGyro~\cite{mustaniemi2019deepgyro}  & 23.78 & 0.6649 & & 5.64 & 0.381 & 31.03      & 769      & 0.068    \\
                           & EggNet~\cite{ji2021eggnet}               & 25.49 & 0.7266 & & 5.18 & 0.413 & 6.34  & 1102   & 0.071    \\
                           & INformer~\cite{ren2024informer}          & 25.11 & 0.7103 & & 5.29 & 0.408 & 24.88 & 1315   & 0.464    \\
                           & Ours                                     & 27.28 & 0.7803 & & 4.47 & 0.548 & 16.31 & 262    & 0.130    \\
\bottomrule[1.5pt]
\end{tabular}
}
\caption{
Quantitative comparison on the \SynthDataName{} and \RealDataName{}-S.
For the \RealDataName{}-S, we use non-reference metrics (NIQE~\cite{mittal2012niqe} and TOPIQ~\cite{chen2024topiq} trained on KonIQ-10k dataset~\cite{hosu2020koniq}) for evaluation.
Inference times were measured using a $720 \times 1280$ image. 
For models utilizing gyro data, inference times include both gyro data embedding construction and the deblurring process.
}
\label{table:quantitative_synth}
\end{table*}

\begin{figure*}[ht]
\centering
\includegraphics[width=\linewidth]{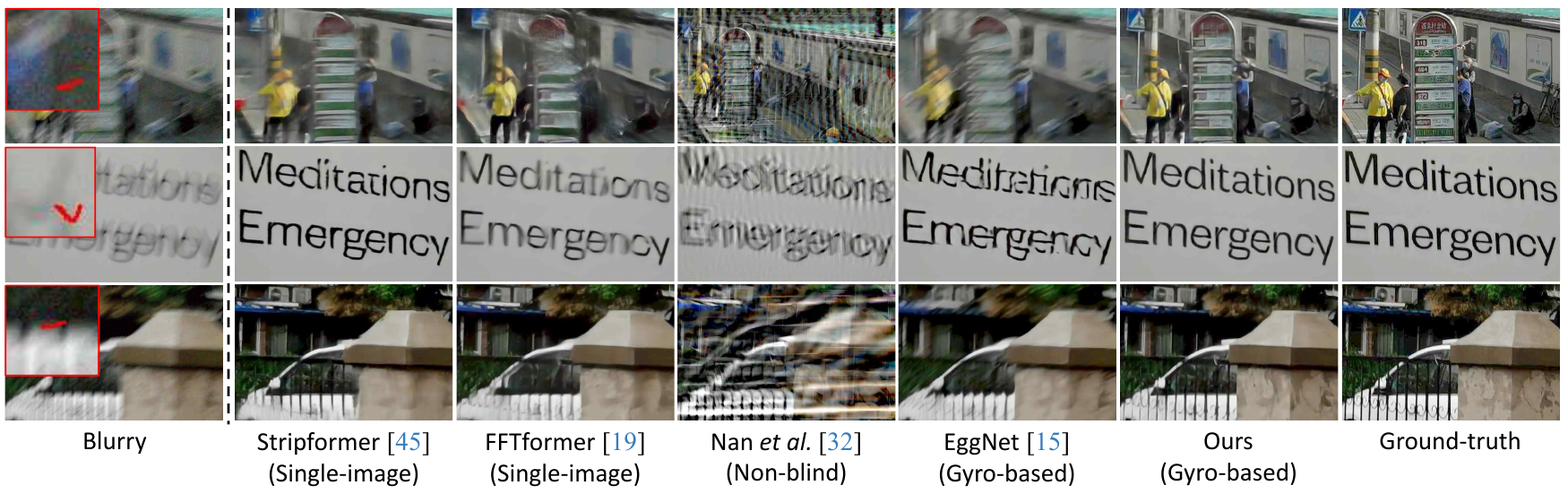}
\caption{Qualitative comparison on \SynthDataName{}.
The red curves overlaid on the blurred images visualize the input gyro data.
}
\vspace{-1mm}
\label{fig:qualitative_synth}
\end{figure*}

\subsection{\RealDataName}


\RealDataName{} consists of two subsets: \RealDataName{}-S and \RealDataName{}-H.
\RealDataName{}-S was collected using a Samsung Galaxy S22, which was also used for constructing the \SynthDataName{} dataset.
We use \RealDataName{}-S as the primary dataset to evaluate the performance of our approach on real-world images.
On the other hand, \RealDataName{}-H was collected using a Huawei P30 Pro, and is utilized for evaluating the generalization ability of our approach.
\RealDataName{}-S provides 100 real-world blurry images of size $4080\times3060$ along with their corresponding gyro data. \RealDataName{}-H provides 17 real-world blurry images of size $5120\times3840$ and their gyro data.
The images in both datasets are provided in DNG and JPEG.

\section{Experiments}
\label{sec:experiments}

We implemented our network using PyTorch.
We trained our network on randomly cropped $256\times256$ image patches from the \SynthDataName{} dataset for 300 epochs with a batch size of 16. 
We used the Adam optimizer~\cite{kingma2014adam} with $\beta_1 = 0.9$ and $\beta_2 = 0.999$. 
The learning rate was initially set to 0.0001 and reduced to 1e-7 using the cosine annealing scheduler~\cite{loshchilov2016cosineannealing}.
We used the PSNR loss~\cite{chen2021hinet, chen2022nafnet} to train our model.
We measured the computation times of all the models on a GeForce RTX 3090 GPU.


We evaluate the performance of our method using the test set of \SynthDataName{} and \RealDataName{}.
Additional experiments and examples including visualization of the effect of gyro refinement and an analysis on the robustness to gyro errors are included in the supplementary material.

\begin{figure*}[t]
\centering
\includegraphics[width=\linewidth]{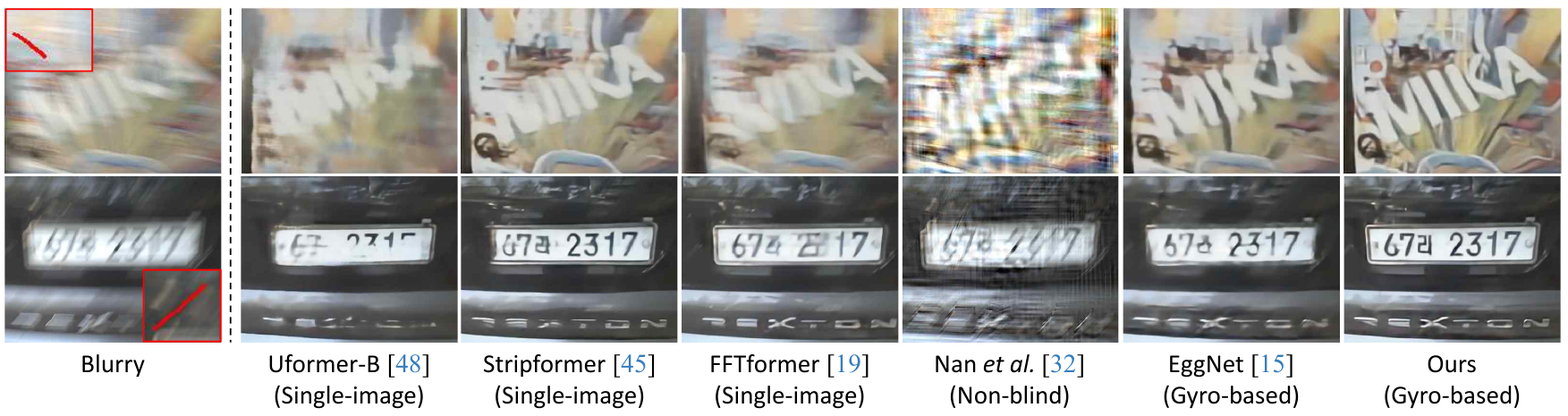}
\caption{Qualitative comparison on \RealDataName{}-S.
The red curves overlaid on the blurred images visualize the input gyro data.
}
\vspace{-1mm}
\label{fig:qualitative_real}
\end{figure*}

\paragraph{Comparison with state-of-the-art methods}
We compare \MethodName{} with state-of-the-art single image deblurring methods that use gyro data~\cite{mustaniemi2019deepgyro, ji2021eggnet, ren2024informer} and those that do not~\cite{zamir2021mprnet, chen2022nafnet, wang2022uformer, Tsai2022Stripformer, kong2023fftformer, mao2024adarevd}.
We also compare \MethodName{} with non-blind deconvolution methods that are designed to handle kernel error~\cite{nan2020nonblinderror, vasu2018nonblinderror}.
Since these methods assume uniform blur kernels, we extended them to handle spatially-varying blur kernels computed from gyro data by using patch-wise blur kernels~\cite{harmeling2010space}.
For more details about our extensions of the non-blind deconvolution methods, we refer the readers to the supplementary material.
We train all models with the training set of \SynthDataName{} and evaluate the performance of all the models on the test set of \SynthDataName{} and \RealDataName{}-S.

The results are presented in \cref{table:quantitative_synth}.
As shown in the table, \MethodName{} significantly outperforms all previous non-gyro-based methods. 
Specifically, compared to transformer-based approaches~\cite{wang2022uformer,Tsai2022Stripformer,kong2023fftformer}, our method requires substantially less computation and shorter computation time, while achieving more than 1 dB higher PSNR. 
Furthermore, our method surpasses all previous gyro-based methods by a large margin, despite requiring a comparable model size and much smaller computational overhead. 
The non-blind deblurring methods designed to handle blur kernel errors also perform significantly worse than our method due to large gyro errors. 
Nonetheless, our approach demonstrates superior performance even in the presence of large gyro errors.

Qualitative comparisons on \SynthDataName{} and \RealDataName{}-S are shown in \cref{fig:qualitative_synth} and in \cref{fig:qualitative_real}, respectively.
As shown in the figures, the previous non-gyro-based methods fail to restore sharp details due to severe blur.
Moreover, despite the input gyro data, the previous gyro-based methods also fail to produce sharp details as they are not robust to gyro errors.
The non-blind deblurring methods that are designed to handle kernel errors show significant ringing artifacts due to large errors in the gyro data.
On the other hand, our method successfully restores sharp images as our method can effectively exploit erroneous gyro data.

\begin{table}[t]
    \centering
\scalebox{0.67}{\begin{tabular}{lccc}
\toprule[1.5pt]
\multirow{2}{*}{{\centering Model}} & \SynthDataName{} & & \RealDataName{}-S \\ \cline{2-2}\cline{4-4}
                                    & \small{PSNR $\uparrow$ / SSIM $\uparrow$} & & \small{NIQE $\downarrow$ / TOPIQ $\uparrow$} \\ \midrule
(a) Deblurring w/ no gyro data              & 24.90 / 0.6997          & & 5.23 / 0.391 \\
(b) Training w/ error-free gyro data        & 24.94 / 0.7112          & & 5.27 / 0.396 \\
(c) Deblurring w/ gyro data w/o any refine. & 25.47 / 0.7128          & & 5.01 / 0.418 \\
(d) Gyro refine.~w/o image features         & 26.17 / 0.7465          & & 4.75 / 0.447 \\ 
(e) Deform.~conv.~using only gyro features  & 26.32 / 0.7542          & & 4.76 / 0.460 \\
(f) \MethodName{}                           & 26.94 / 0.7667          & & 4.61 / 0.511 \\ \midrule
(g) \MethodName{} + curriculum learning     & 27.28 / 0.7803          & & 4.47 / 0.548 \\ \bottomrule[1.5pt]
\end{tabular}}
\caption{Ablation study on the components of \MethodName{}. For the \RealDataName{}-S, we use non-reference metrics (NIQE~\cite{mittal2012niqe} and TOPIQ~\cite{chen2024topiq} trained on KonIQ-10K dataset~\cite{hosu2020koniq}) for evaluation.}
\vspace{-3mm}
\label{table:network_ablation}
\end{table}

\begin{figure}[t]
\includegraphics[width=\linewidth]{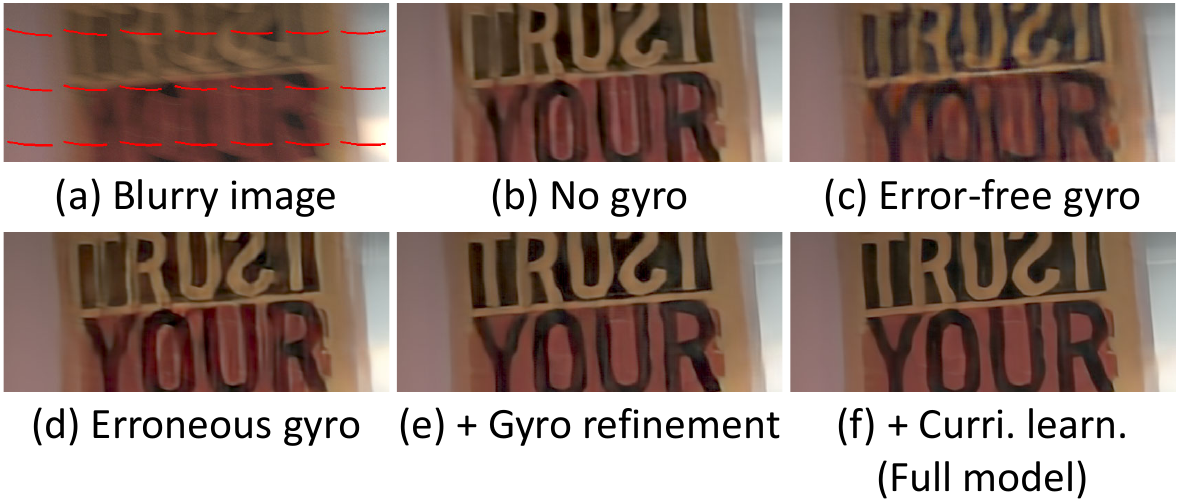}
\centering
\vspace{-6mm}
\caption{Qualitative results of the ablation study on a real-world blurry image.
}
\vspace{-3mm}
\label{fig:ablation_qual}
\end{figure}

\paragraph{Ablation study} We conducted an ablation study to investigate the effects of individual components in our method. 
To this end, we built several variants of \MethodName{} and compared their performances.
Both quantitative and qualitative results are reported in \cref{table:network_ablation} and \cref{fig:ablation_qual}, respectively. 



\cref{table:network_ablation}-(a) corresponds to a model with no gyro modules, using only image features in the gyro deblurring blocks.
\cref{table:network_ablation}-(b) represents our full model, but trained with error-free gyro data.
\cref{table:network_ablation}-(c) is a model that is trained with erroneous gyro data but does not refine gyro features using image features. Specifically, in this model, gyro features are not concatenated with image features in both gyro refinement and gyro deblurring blocks. 
\cref{table:network_ablation}-(f) is our full model trained without curriculum learning, while \cref{table:network_ablation}-(g) is our full model trained with curriculum learning.

As shown in \cref{table:network_ablation}-(a), deblurring with no gyro data performs the worst, as it cannot benefit from the blur cue provided by gyro data. 
The model in (b) achieves only marginal improvement over (a), indicating the importance of reflecting gyro errors in training data.
The results in (c), (d), (e) and (f) demonstrate the necessity of training with erroneous gyro data and gyro refinement using image features.
Finally, curriculum learning in (g) further improves deblurring quality, as it helps the model to fully utilize erroneous gyro data.


\cref{fig:ablation_qual} also highlights the impact of each component in our method. 
The results in \cref{fig:ablation_qual}-(b) and (c) reveal that training with error-free gyro data yields no performance gain over training without gyro data since error-free gyro data cannot reflect real-world gyro errors.
In contrast, \cref{fig:ablation_qual}-(d) shows that training with erroneous gyro data aids in handling these errors but remains suboptimal. 
Gyro refinement using image features in \cref{fig:ablation_qual}-(e) significantly improves image quality, while \cref{fig:ablation_qual}-(f) demonstrates how curriculum learning further enhances error handling.

\begin{figure}[t!]
\includegraphics[width=\linewidth]{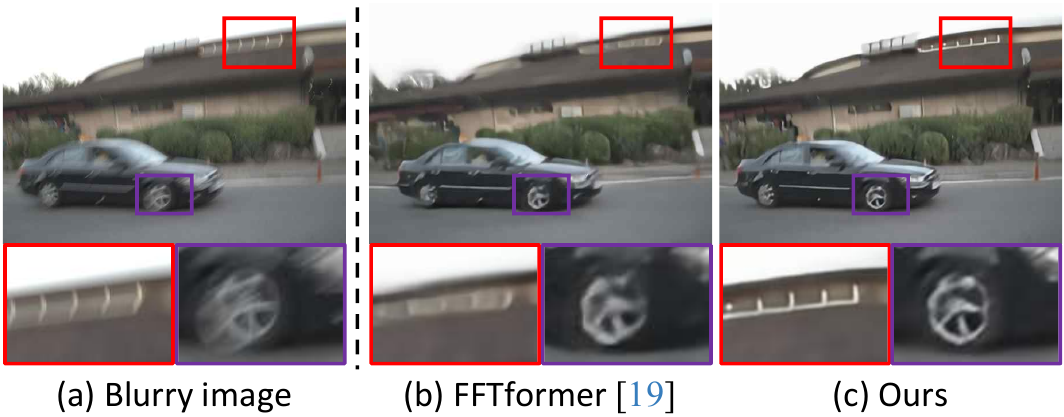}
\centering
\caption{Results of a real-world blurred image with a moving object. (a) Blurry image and its background region and foreground object. (b) Result of FFTformer. (c) Result of \MethodName{}.}
\vspace{-1mm}
\label{fig:moving_objects}
\end{figure}

\paragraph{Moving objects}
Another source of gyro error is moving objects with different blur trajectories.
\cref{fig:moving_objects} shows a real-world example where the input blurry image has a moving object with a different blur trajectory.
Due to severe blur, the result of FFTformer~\cite{kong2023fftformer}, the state-of-the-art non-gyro-based method, still has remaining blur in the background. 
In contrast, our method successfully restores sharp details on the background and shows comparable quality on the moving object thanks to its robustness to gyro errors.

\begin{figure}[t!]
\includegraphics[width=\linewidth]{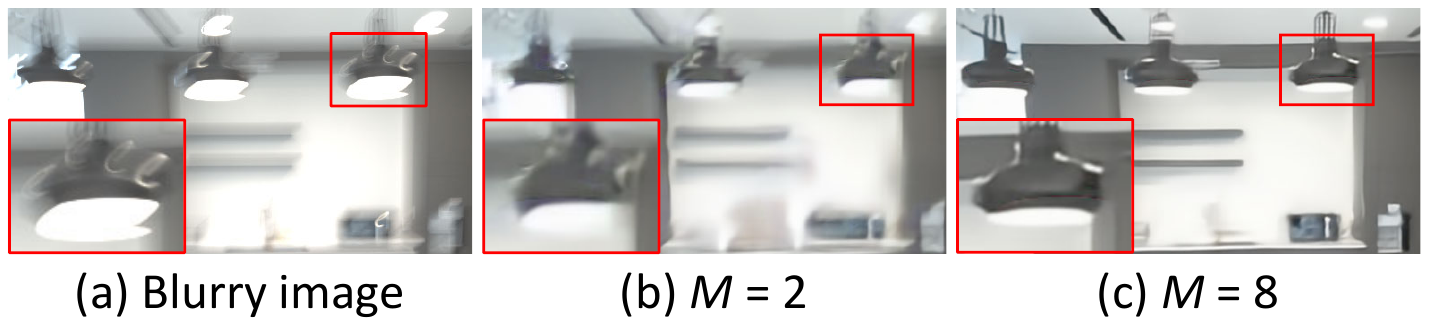}
\centering
\caption{Effect of increasing $M$ for real-world complex motion blur.
(a) Real-world blurry image with complex blur. (b) Result with $M = 2$. (c) Result with $M = 8$.
}
\vspace{-1mm}
\label{fig:complex_motion}
\end{figure}

\begin{table}[t]
\centering
    \scalebox{0.9}{
        \begin{tabular}{c|cccc}
            \Xhline{4\arrayrulewidth}
            $M$  & 2      & 4      & 8      & 16     \\ \hline\hline
            PSNR & 25.71  & 26.80  & 27.28  & 27.32  \\
            SSIM & 0.7706 & 0.7651 & 0.7803 & 0.7811 \\ \Xhline{4\arrayrulewidth}
        \end{tabular}
    }
\caption{Effect of \GyroReprName{} hyperparameter $M$.}
\vspace{-1mm}
\label{table:dmgv_ablation}
\end{table}

\paragraph{Camera motion field hyperparameter \boldmath$M$}
We investigate the impact of the hyperparameter $M$ that determines the temporal resolution of the \GyroReprName{}.
As shown in \cref{table:dmgv_ablation}, the deblurring quality improves gradually as $M$ increases.
This highlights the importance of gyro data representation that can capture real-world intricate camera motions, and also explains the limited performance of previous gyro-based approaches~\cite{ji2021eggnet,mustaniemi2019deepgyro}.
\cref{fig:complex_motion} also demonstrates the effect of increasing the number of vectors to handle complex real-world motion blur.
However, increasing $M$ to 16 does not yield a significant performance gain while it doubles the memory consumption, indicating that $M=8$ is sufficient to represent the camera shakes in most of the images in the test set of \SynthDataName{}.

\begin{figure}[t]
\centering
\includegraphics[width=\linewidth]{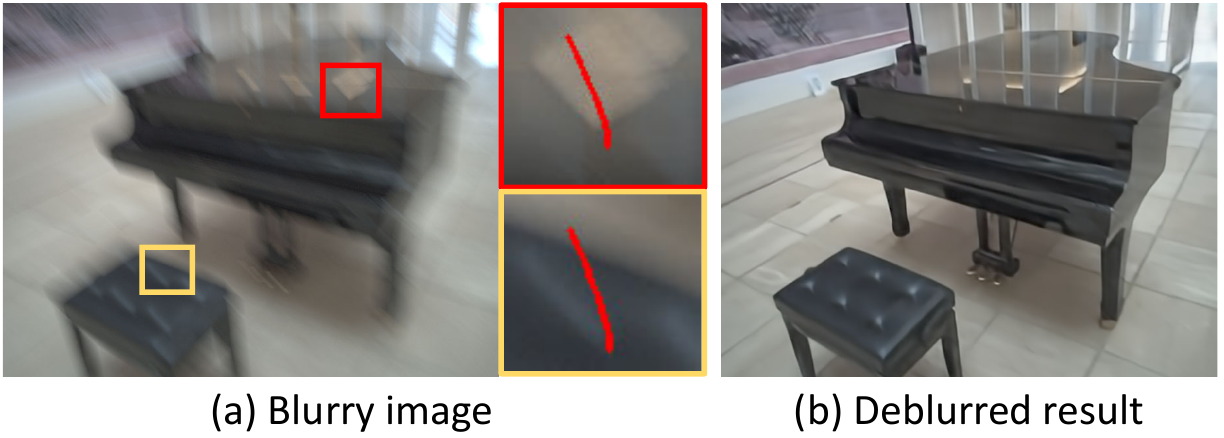}
\caption{Blurry image taken with a Huawei P30 Pro device and its deblurred result. (a) A blurry image and its gyro errors. (b) Deblurred result of (a).
}
\vspace{-1mm}
\label{fig:huawei}
\end{figure}

\paragraph{Generalization to other devices}
We test \MethodName{}'s generalization ability by comparing it with top deblurring methods from \cref{table:quantitative_synth} on \RealDataName{}-H. 
All methods were trained on \SynthDataName{}. 
The results in \cref{table:huawei} show our approach outperforms the others, even on a different device.
\cref{fig:huawei} also demonstrates that our approach successfully removes blur from an image captured by a different device, highlighting its generalization ability.

\begin{table}[t]
\centering
    \scalebox{0.79}{
        \begin{tabular}{c|cccc}
            \Xhline{4\arrayrulewidth}
            Model & Stripformer~\cite{Tsai2022Stripformer} & FFTformer~\cite{kong2023fftformer} & EggNet~\cite{ji2021eggnet} & Ours     \\ \hline\hline
            NIQE $\downarrow$ & 5.47  & 5.22  & 5.71  & 4.94  \\
            TOPIQ $\uparrow$  & 0.451 & 0.459 & 0.432 & 0.521 \\ \Xhline{4\arrayrulewidth}
        \end{tabular}
    }
\caption{Quantitative evaluation on \RealDataName{}-H using non-reference metrics~\cite{mittal2012niqe, chen2024topiq}.}
\vspace{-3mm}
    \label{table:huawei}
\end{table}

\section{Conclusion}
\label{sec:conclusion}

In this paper, we proposed \MethodName{}, a novel gyro-based single image deblurring method that can effectively restore sharp images with the help of gyro data.
To fully exploit gyro data while considering real-world gyro error, we presented a novel gyro refinement block, a novel gyro deblurring block, and a curriculum learning-based training strategy.
In addition, we also presented the \GyroReprName{}, a novel gyro embedding scheme to represent real-world camera motions.
Finally, we proposed synthetic and real datasets where blurred images are paired with gyro data.
Extensive quantitative and qualitative experiments demonstrate the effectiveness of our approach.

\paragraph{Limitations and future work}
Our method has a few limitations.
While our method uses a fixed value for the hyperparameter $M$, the complexity of camera shakes may increase along with the exposure time, which means that a longer exposure time may require a larger value for $M$.
Our method does not use the accelerometer, which is also equipped with most smartphones, and which may provide additional valuable information on the camera motion.
Our image deblurring module adopts a relatively simple network architecture compared to recent non-gyro-based single image deblurring approaches, and this may limit the deblurring performance of our method.
Extending our work to address these limitations would be an interesting future direction.

\paragraph{Acknowledgements}
This work was supported by the National Research Foundation of Korea (NRF) grants (RS-2023-NR077065, RS-2023-00280400, RS-2023-00211658, RS-2024-00438532) and by Institute of Information \& communications Technology Planning \& Evaluation (IITP) grant (RS-2019-II191906, Artificial Intelligence Graduate School Program (POSTECH)) funded by the Korea government (MSIT). 
This work was also supported by Samsung Research Funding \& Incubation Center of Samsung Electronics under Project Number SRFC-IT1801-52.

{
    \small
    \bibliographystyle{ieeenat_fullname}
    \bibliography{main}
}


\end{document}